\documentclass[11pt,a4paper]{article}
\usepackage[nohyperref]{naaclhlt2019}
\usepackage{latexsym}

\usepackage{naaclhlt2019}  
\usepackage{times}  
\usepackage{helvet}  
\usepackage{courier}  
\usepackage{url}  
\usepackage{graphicx}  
\usepackage{amsmath}
\usepackage{amssymb}
\usepackage{enumitem}
\usepackage{graphicx}
\usepackage{subcaption}
\usepackage{color,soul}
\usepackage{algorithm}
\usepackage[noend]{algpseudocode}

\DeclareMathOperator*{\argmax}{arg\,max}  

\aclfinalcopy

\begin{document}

\title{Detecting cognitive impairments by agreeing on interpretations \\of linguistic features}

\author{Zining Zhu$^{1,2}$, Jekaterina Novikova$^{1}$, Frank Rudzicz$^{3,5,2,4,1}$ \\
  $^1$: Winterlight Labs 
  $^2$: University of Toronto \\
  $^3$: Li Ka Shing Knowledge Institute, St Michael's Hospital, Toronto ON, Canada\\
  $^4$: Vector Institute for Artificial Intelligence 
  $^5$: Surgical Safety Technologies\\
  {\tt \{zining, jekaterina\}@winterlightlabs.com, frank@cs.toronto.edu} \\}
 
\maketitle
\begin{abstract}
Linguistic features have shown promising applications for detecting various cognitive impairments. To improve detection accuracies, increasing the amount of data or the number of linguistic features have been two applicable approaches. However, acquiring additional clinical data can be expensive, and hand-crafting features is burdensome. In this paper, we take a third approach, proposing Consensus Networks (CNs), a framework to classify after reaching agreements between modalities. We divide linguistic features into non-overlapping subsets according to their modalities, and let neural networks learn low-dimensional representations that agree with each other. These representations are passed into a classifier network. All neural networks are optimized iteratively.

In this paper, we also present two methods that improve the performance of CNs. We then present ablation studies to illustrate the effectiveness of modality division. To understand further what happens in CNs, we visualize the representations during training. Overall, using all of the 413 linguistic features, our models significantly outperform traditional classifiers, which are used by the state-of-the-art papers. 

\end{abstract}
\section{Introduction}
Alzheimer's disease (AD) and its usual precursor, mild cognitive impairment (MCI), are prevalent neurodegerative conditions that inhibit cognitive abilities. Cognitive impairments are traditionally diagnosed only with standard clinical tests like MoCA \citep{MoCA} and the Rey-Auditory Verbal learning Test \citep{Rey1941}, but hiring clinicians to administer these tests and analyze their results is costly. 
Fortunately, many cognitive impairments can be observable in daily life, because they impact one's language abilities.
For example, cognitively impaired people tend to use more pronouns instead of nouns, and pause more often between sentences in narrative speech \citep{Roark2011}. 

This insight makes automatic detection possible. Machine learning classifiers can detect cognitive impairments given descriptive linguistic features.
In recent work, linguistic features including pronoun-noun-ratios, pauses, and so on, are used to train classifiers to detect cognitive diseases in various tasks. For example, \citet{fraser15-JAD} achieved up to 82\% accuracy on DementiaBank\footnote{\url{https://talkbank.org/DementiaBank}}, the largest publicly available dataset on detecting cognitive impairments from speech, and \citet{weissenbacher2016automatic} achieved up to 86\% accuracy on a corpus of 500 subjects. \citet{yancheva-fraser-rudzicz:SLPAT2015} estimated Mini-Mental State Estimation scores (MMSEs), describing the cognitive status and characterizing the extent of cognitive impairment.

To improve the accuracy of automated assessment using engineered linguistic features, there are usually two approaches: incorporating more clinical data or calculating more features. 
Taking the first approach, \citet{noorian2017importance} incorporated normative data from Talk2Me\footnote{\url{https://www.cs.toronto.edu/talk2me/}} and the Wisconsin Longitudinal Study \citep{herd2014WLS} in addition to DementiaBank, which increased AD:control accuracy up to 93\%, and moderateAD:mildAD:control three-way classification accuracy to 70\%. Taking the second approach, \citet{yancheva2016vector} used 12 features derived from vector space models and reached a .80 F-score on DementiaBank. \citet{santos2017enriching} calculated features depicting characteristics of co-occurrence graphs of narrative transcripts (e.g., the degree of each vertex in the graph). Their classifiers reached 65\% accuracy on DementiaBank (MCI versus a subset of Control).

There are limitations in either of the two approaches. On one hand, acquiring additional clinical data can be expensive \citep{berndt2013price}. Moreover, the additional data should be similar enough to existing training data to be helpful.
On the other hand, crafting new features requires creativity and collaboration with subject matter experts, and the implementation can be time consuming. Neither of these approaches is satisfactory.

These limitations motivate us to take a third, novel approach. Instead of using new data or computing new features, we use the existing linguistic features.

If the speaker is cognitively impaired, and their language ability is affected, features from each of the acoustic, syntactic, and semantic modalities should reflect such change \citep{Szatloczki2015,Moro2015schizophrenia,fraser15-JAD}. We therefore need to distill the common information revealed by features from multiple, mainly distinct, modalities.

To utilize information common across different modalities, \citet{becker1992self} and \citet{de1994learning} let classifiers look at each modality and supervise each other. These examples illustrated the effectiveness of multi-view learning in utilizing common information among different observations, but their algorithms fail to train useful classifiers for cognitive impairments in our datasets. Without explicit supervision, self-supervised models almost always converge to a state producing the same predictions for all people, giving trivial classifiers.  

Instead of aligning the predictions from modalities, we let the representations of the modalities agree. Generative adversarial networks (GANs) provide an approach. In GANs, a ``discriminator'' network is trained to tell whether a vector is drawn from the real world or produced synthetically by a ``generator'' neural network, while the generator is trained to synthesize images as close to real data as possible.
We borrow this setting, and encourage the neural networks interpreting different modalities to produce representations of modalities as similar to each other as possible. This leads to our classifier framework, consensus networks (CNs).

Consensus networks constitute a framework using adversarial training to utilize common information among modalities for classification. In this framework, several neural networks (``ePhysicians'') are juxtaposed, each learning the representation of a partition of linguistic features for each transcript. Being trained towards producing agreed representations, we show they are increasingly able to capture common information contained across disparate subsets of linguistic features. 

We empirically add two extensions to CN that improve the classification accuracies, called the ``noise modality'' and ``cooperative optimization'' respectively, as explained below.
To illustrate the effectiveness of the consensus mechanisms, we present two ablation studies. First, we compare neural networks built with consensus (CN) and those without (MLP). On partial or complete modalities, CN outperforms MLP significantly. Second, we compare CNs built with linguistic features divided into random subsets. Division according to their natural modalities train better consensus networks. We also visualize the representations during training procedure, and show that when the representations agree, their distributions appear symmetric.

Overall, taking all 413 linguistic features, our models significantly outperform traditional classifiers (e.g., support vector machines, quadratic discriminant analysis, random forest, Gaussian process), which are used by the state-of-the-art.

\section{Related Works}
\paragraph{Generative Adversarial Networks}
The idea of aligning representations by making them indistinguishable is inspired by GAN \citep{goodfellow2014generative}, where a generator produces fake images (or other data) that are as similar to real data as possible. However, our model does not have a generator component as GANs do. Instead, we only compress features into representations while trying to align them.

\paragraph{Multi-view learning}
Learning from multiple modalities is also referred to as multi-view learning. \citet{becker1992self} set up multiple neural networks to look at separate parts of random-dot stereograms of curved surfaces, and urge their prediction to equal each other. The trained neural networks were able to discover depth without prior knowledge about the third dimension. 
\citet{de1994learning} divided linguistic features into two modalities, and passed them to two separate neural networks. The two neural networks supervised each other (i.e., output labels that are used to train the other) during alternative optimization steps to reach a consensus. Their self-supervised system reached 79$\pm 2\%$ accuracy using the Peterson-Barney vowel recognition dataset \citep{peterson1952control}.
\citet{benediktsson1997parallel} computed multiple views from the same feature sets and classified by taking their majority votes. \citet{Pou-Prom2018} used canonical correlation analysis (CCA) to classify using multiple aspects. Contrary to that work, our consensus networks take in distinct subsets of features as modalities. 
Co-training \citep{blum1998co-training} and tri-training \citep{zhou2005tri} use distinct subsets of features, but they use them to train distinct classifiers, and let the results directly supervise each other. Their approach `bootstrapped' classifications based on a few labeled data, but our method explicitly uses a modality discriminator that enforces alignments between modalities.

\paragraph{Domain Adaptation}
In domain adaptation and multi-task learning, there have been many attempts to learn indistinguishable embeddings between domains. For example, \citet{Ganin2016} and \citet{Joty2017} applied a gradient reversal layer to let encoders minimize the domain classification loss. 
\citet{Baktashmotlagh2013} minimized the maximum-mean discrepancy (MMD) loss in a reproductive kernel Hilbert space (RKHS) of the latent representations.
\citet{Motiian2017} used semantic similarity loss between latent representations of different class data to encourage alignments between domains.
\citet{liu2017adversarial} and \citet{Chen2018} used shared and private networks to learn information contained either commonly in domains or domain-specific. 
Our work is unique. First, there is only one domain in our problem setting. Second, we use iterative optimization to encourage discrepancies between domains. Third, we have two empirical improvements (noise modality and cooperative optimization) that make our Consensus Networks outperform traditional classifiers.

\section{Methods}
\subsection{Dataset}
We use DementiaBank, the largest publicly available dataset for detecting cognitive impairments. It includes verbal descriptions (and associated transcripts) of the Cookie Theft picture description task from the Boston Diagnostic Aphasia Examination \cite{PittsCorpus}. The version we have access to contains 240 speech samples labeled Control (from 98 people), 234 with AD (from 148 people), and 43 with MCI (from 19 people)\footnote{The version of DementiaBank dataset we acquired contains a slightly different number of samples from what some previous works used. In Control:AD, \citet{fraser15-JAD} used 233 Control and 240 AD samples; \citet{yancheva2016vector} had 241 Control and 255 AD samples; \citet{hernandez2018computer} had 242 Control and 257 AD samples (with 10\% control samples excluded from the evaluation). In Control:MCI, \citet{santos2017enriching} used all 43 transcriptions from MCI and 43 sampled from the Control group. With no clear descriptions of the sampling procedures, the constituents of the Control group might differ from our sample. In this paper, we run our models on the same tasks (i.e., Control:AD) and compare to the results of models used in the literature.}. All participants were older than 44 years.

\subsection{Linguistic features}
The dataset contains narrative speech descriptions and their transcriptions. We preprocess them by extracting 413 linguistic features for each speech sample. These linguistic features are proposed by and identified as the most indicative of cognitive impairments by various previous works, including \citet{roark2007syntactic}, \citet{chae2009predicting}, \citet{Roark2011}, \citet{fraser15-JAD}, and \citet{hernandez2018computer}. 
After calculating these features, we use KNN imputation to replace the undefined values (resulting from divide-by-zero, for example), and then normalize the features by their $z$-scores.
The following are brief descriptions of these features, grouped by their natural categories. More detailed descriptions are included in the Appendix.

There are 185 acoustic features (e.g., average pause time), 117 syntactic features (e.g., Yngve statistics \citep{yngve1960model} of the parse tree, computed by the LexParser in CoreNLP \citep{corenlp}), and 31 semantic features (e.g., cosine similarity between pairs of utterances) Moreover, we use 80 part-of-speech features that relate to both syntax and semantics but are here primarily associated with the latter.




\paragraph{Modality division}
After representing each sample with a 413-dimensional vector $\mathbf{x}$ consisting of all available linguistic features, we divide the vector into $M$ partitions (`modalities') of approximately equal sizes [$\mathbf{x_1}$, $\mathbf{x_2}$, ..., $\mathbf{x_M}$], according to the groups mentioned above. Unless mentioned otherwise (e.g., in the ablation study shuffling modalities), this is our default choice for assigning modalities.

\begin{figure}[t] 
\centering
\includegraphics[width=\linewidth]{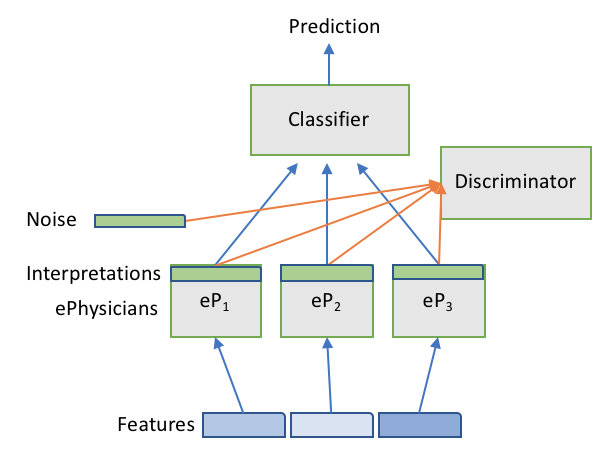}
\caption{Overview of model structure when features (blue rectangles) are divided into three modalities (non-overlapping subsets). Each subset of features are passed into an ``ePhysician" neural network whose outputs (green rectangles) are the representations. They are passed (one by one) into a ``Discriminator" neural network and (after combined) into a ``Classifier" network, respectively.}
\label{fig:model-structure}
\end{figure}

\subsection{Model}
Figure \ref{fig:model-structure} is an example of our model structure (with $M=3$ modalities), and this section elaborates the inference procedure, the training algorithm, and our two improvements.

\paragraph{Inference}
With the extracted linguistic features divided into subsets by their modalities, each speech sample is described by $M=3$ non-overlapping feature vectors $\mathbf{x} = [\mathbf{x_1}, ..., \mathbf{x_m} ]$. 
These feature vectors are then passed into corresponding ePhysician networks, each outputting a vector $\mathbf{i_m}$, which is a distilled representation of the subject from a modal-specific perspective (e.g., the semantic). We also refer to it as the \emph{interpretation vector} and use them interchangeably. Formally, the $m^{th}$ ePhysician can be written as a function, $f_m(.)$ generating the representation:
$$\mathbf{i_{m}} = f_{m}(\mathbf{x_m})$$

To challenge the similarity of representations from different modalities, we let a discriminator neural network $f_D(.)$ take in each of the $M$ representations and predict the likelihood of the originating modality $m$.

$$P(m=k\,|\,\mathbf{i}) = \frac{e^{f_D(\mathbf{i})_k}}{\sum_k e^{f_D{(\mathbf{i})}_k}}$$
where $k=1,...,M$.

To attempt a diagnosis, a classifier network $f_C(.)$ takes in the combination of $M$ representations of each speech sample, and outputs a prediction probability for detection result y:
$$P(y=l\,|\,\mathbf{x}) = \frac{e^{f_C(\mathbf{i_{1..M}})_l}}{\sum_l e^{f_C(\mathbf{i_{1..M}})_l}}$$  
where $l\in \{0,1\}$ for two-class classification (i.e., 0 for healthy and 1 for dementia).
The predicted class corresponds to those with the highest probability:
$$\hat{y} = \argmax_l P(y=l|\mathbf{x})$$

\paragraph{Optimization} The training procedure optimizes the adversarial objective, and the conventional classifier objective: 
\begin{itemize}
\item The adversarial objective sets up the ePhysicians and the Discriminator to work in an adversarial manners. The ePhysicians try to produce indistinguishable representations, while the discriminator tries to tell them apart.  
\begin{equation}\begin{aligned}
\label{eq:first-goal}
&\min_{D} \max_{P} \mathcal{L_D} \text{ where }  \\
\mathcal{L_D} &= \mathbb{E}_{\mathbf{x}} \mathbb{E}_{m=1..M} \left\{-\text{log} P(\hat{m}=m|\mathbf{i_m}) \right\}
\end{aligned}\end{equation}

\item Make the classifier network as accurate as possible. This is done by minimizing the cross entropy classification loss: 
\begin{equation}\begin{aligned}
\label{eq:second-goal}
&\min_{C} \mathcal{L_C} \text{ where } \\
\mathcal{L_C} &= \mathbb{E}_{\mathbf{x}} \left\{ -\text{log} P(\hat{y}=y|\mathbf{i_{1..M}}) \right\}
\end{aligned}\end{equation}
\end{itemize}

Overall, $\displaystyle \min_C \mathcal{L_C}$ and $\displaystyle \min_D \max_P \mathcal{L_D}$ set up a complex optimization problem. We use iterative optimization steps, similar to \citet{goodfellow2014generative}.

There are two tricks that we found to improve the performance of the models. Namely, the \emph{noise modality} and the \emph{cooperative optimization}. We explain them below.

\paragraph{Noise modality} For each participant session, we add a ``noise modality representation" $\mathbf{i_0}$ drawn from a Gaussian distribution with the mean and variance identical to those of other representation vectors. 
$$\mathbf{i_0} \sim \mathcal{N}(\mu_{\mathbf{i_{1..M}}}, \sigma^2_{\mathbf{i_{1..M}}})$$
This additional representation vector is passed into the discriminator, but not passed into the classifier. The first optimization goal (\ref{eq:first-goal}) is therefore:
\begin{equation}\begin{aligned}
\label{eq:modified-first-goal}
&\min_{D} \max_{P} \mathcal{L_D} \\
\text{ where } \mathcal{L_D} &= \mathbb{E}_{\mathbf{x}}\mathbb{E}_{m=0..M} \left\{-\text{log} P(\hat{m}=m|\mathbf{i_m}) \right\}
\end{aligned}\end{equation}
To some extent, the noise representation vector works like a regularization mechanism to refrain the discriminator from making decisions based on superficial statistics. We show in \ref{subsec:noise-mod} that this addition empirically improves classifier performance.

\paragraph{Cooperative optimization} When optimizing the classifier, we find that allowing gradients to propagate back to the ePhysicians improves the model's overall performance. During optimization, the ePhysicians need to \emph{cooperate} with the classifier (while adversarial to the discriminator). The second optimization goal (\ref{eq:second-goal}) is therefore:
\begin{equation}\begin{aligned}
\label{eq:modified-second-goal}
&\min_{C,P} \mathcal{L_C}\\
\text{ where } \mathcal{L_C} &= \mathbb{E}_{\mathbf{x}} \left\{ -\text{log} P(\hat{y}=y|\mathbf{i_{1..M}}) \right\}
\end{aligned}\end{equation}

\paragraph{Implementation} As a note of implementation, all ePhysicians, the classifier, and the discriminator networks are fully connected networks with Leaky ReLU activations \cite{nair2010rectified} and batch normalization \cite{ioffe2015batch}. The hidden layer sizes are all 10 for all ePhysician networks, and there are no hidden layers for the discriminator or classifier networks. Although modalities might contain slightly different numbers of input dimensions, we do not scale the ePhysician sizes. This choice comes from the intuition that the ePhysicians should extract into the representation as similar information as possible. We use three Adam optimizers \cite{kingma2014adam}, each corresponding to the minimization of ePhysician, Discriminator, and the Classifier, and optimize iteratively for no more than 100 steps. The optimization is stopped prior to step 100 if the classification loss $\mathcal{L_C}$ converges (i.e., does not differ from the previous iteration by more than $1\times 10^{-4}$) on training set. The train / validation / test set are divided randomly in 60/20/20 proportions.

\begin{algorithm}[H]  
  \caption{The overall algorithm}\label{alg:age-indep-simple}
  \begin{algorithmic}[1]
  \State Initialize the networks
  \For {step := 1 to N} \Comment{N is a hyper-param}
    \For{minibatch $\mathbf{x}$ in training data $\mathcal{X}$}
      \For{modality m := 1 to M} 
        \State $\mathbf{i_m} = I_m(\mathbf{x_m})$
      \EndFor
      \State Sample the noise modality $\mathbf{i_0}$
      \State Calculate $\mathcal{L_D}$ with $\mathbf{i_{0..M}}$
      \State Concatenate $\mathbf{i_{1..M}}$ and calculate $\mathcal{L_C}$
      
      \State $\displaystyle \min_{C, P} \mathcal{L_C}$  \Comment{Cooperative optimization}
      \State $\displaystyle \max_{P} \mathcal{L_D}$
      \For{ k:=1 to K} \Comment{K is a hyper-param}
        \State $\displaystyle \min_{D} \mathcal{L_D}$
      \EndFor
    \EndFor
  \EndFor
  \end{algorithmic}
\end{algorithm}

\section{Experiments}
We first show the effectiveness of our two improvements to the model. Next, we do two ablation studies on the arrangements of modalities. Then, we evaluate our model against several traditional supervised learning classifiers used by state-of-the-art works. 
To understand the model further, we also visualize the principal components of the representation vectors throughout several runs.

\subsection{Noise modality improves performance}
\label{subsec:noise-mod}

We compare a CN model with a noise modality to one without (with other hyper parameters including hidden dimensions and learning rates identical). 

Table \ref{tab:noise-mod-exp-comparison} shows that in the AD:MCI classification task, the model with an additional noise modality is better than the one without ($p=0.04$ on 2-tailed $t$-test with 18 DoF). 
Here is a possible explanation. Without adding a noise modality, the discriminators may simply look at the superficial statistics, like the mean and variances of the representations. This strategy tends to neglect the detailed aspects encoded in the representation vectors. Adding in the noise modality penalizes this strategy and trains better discriminators by forcing them to {\em study the details}.

In the following experiments, all models contain the additional noise modality.

\begin{table}[h]
\centering 
\begin{tabular}{|l c c|}
\hline
Model & Micro F1 & Macro F1 \\ \hline 
Noise & \textbf{.7995 $\pm$ .0450} & \textbf{.7998 $\pm$ .0449} \\ 
No noise & .7572 $\pm$ .0461 & .7577 $\pm$ .0456 \\ \hline
\end{tabular}
\caption{Comparison of models with and without representations in noise modality. The models containing a Gaussian noise modality outperform those without.
\label{tab:noise-mod-exp-comparison}}
\end{table}

\subsection{Effectiveness of cooperative optimization}
\label{subsec:c-or-cp}

The second improvement, cooperative optimization, also significantly improves model performance. We compare Consensus Network classifiers trained with cooperative optimization (i.e., $\displaystyle \min_{C,P_{1..M}} \mathcal{L_C}$) to models with the same hyper-parameters but trained non-cooperatively (i.e., $\displaystyle \min_{C} \mathcal{L_C}$). As shown in Table \ref{tab:c-or-cp}, the cooperative variant produces higher-score classifiers than the non-cooperative one ($p<0.001$ on a 2-tailed $t$-test with 18 DoF).
With the cooperative optimization setting, the ePhysicians are encouraged towards producing representations both indistinguishable (by the discriminator) and beneficial (for the classifier). Although the representations might agree less with each other, they could contain more {\em complementary} information, leading to better overall classifier performances.

In other experiments, all of our models use cooperative optimization. 

\begin{table}[h]
\centering 
\begin{tabular}{|l c c|}
\hline 
Optimization & Micro F1 & Macro F1 \\ \hline 
Non-coop & .6696 $\pm$ .0511 & .6743 $\pm$ .0493 \\ 
Cooperative & \textbf{.7995 $\pm$ .0450} & \textbf{.7998 $\pm$ .0449} \\ \hline 
\end{tabular}
\caption{Comparison of models optimized in cooperative and non-cooperative manner.
\label{tab:c-or-cp}}
\end{table}

\subsection{Agreement among modalities is desirable}
In this and the next experiment, we illustrate the effectiveness of our models on different configurations of modalities in an ablation study. We show that our models work well because of the effectiveness of the ``consensus between modalities'' scheme.

In this experiment, we compare our Consensus Network models (i.e., with agreements) with fully-connected neural network classifiers (i.e., without agreements) taking the same partial input features. The networks are all simple multiple layer perceptrons containing the same total number of neurons as the `classifier pipeline' of our models (i.e., ePhysicians plus the classifier)\footnote{For example, for models taking in two modalities, if our model contain ePhysicians with one layer of 20 hidden neurons, the interpretation vector dimension 10, and classifier 5 neurons, then the benchmarking neural network contains three hidden layers with [20$\times$2, 10$\times$2, 5] neurons.} with batch normalization between hidden layers. 
A few observations could be made from Table \ref{tab:fc-mods}:
\begin{enumerate}
	\item Some features from particular modalities are more expressive than others. For example, acoustic features could be used for building better classifiers than those in the semantic ($p=.005$ for 2-tailed $t$-test with 18 DoF) or syntactic modality ($p<.001$ for 2-tailed $t$-test with 18 DoF). More specifically, the syntactic features themselves do not tell much. We think this is because the syntactic features are largely based on the contents of the speech, and remain similar across speakers. For example, almost none of the speakers asked questions, giving zero values in ``occurrences'' of corresponding syntactic patterns.
    \item Our model is able to utilize multiple modalities better than MLP. For MLP classifiers, combining features from different modalities does not always give better models. The syntactic modality features confuse MLP and ``drag down'' the accuracy. However, our models built with the consensus framework are able to utilize the less informative features from additional modalities. In all scenarios using two modalities, our models achieve accuracies higher than neural networks trained on \emph{any} of the two individual modalities.
    \item Given the same combinations of features, letting neural networks produce representation in agreement \emph{does} improve the accuracy in all four scenarios\footnote{$p=3\times 10^{-12}$ on syntactic+semantic features, $p=0.044$ on acoustic + semantic, $p=0.005$ on acoustic + syntactic, and $p=0.046$ on all modalities. All 18 DoF one-tailed $t$-tests.}.
\end{enumerate}

\begin{table}[h]
\centering 
\begin{tabular}{|l c|}
\hline 
Models (Modality) & Accuracy \\ \hline 
MLP (Acoustic) & .7519 $\pm$ .0245  \\
MLP (Syntactic) & .5222 $\pm$ .0180 \\
MLP (Semantic) & .6987 $\pm$ .0278 \\ \hline 
MLP (Syntactic + Semantic) & .5819 $\pm$ .0216 \\
CN (Syntactic + Semantic) & .7257 $\pm$ .0344 \\ \hline 
MLP (Acoustic + Semantic) & .7002 $\pm$ .1128 \\  
CN (Acoustic + Semantic) & .7542 $\pm$ .0433 \\ \hline 
MLP (Acoustic + Syntactic) & .6776 $\pm$ .0952 \\ 
CN (Acoustic + Syntactic) & .7574 $\pm$ .0361 \\ \hline 
MLP (All 3 modalities) & .7528 $\pm$ .0520  \\
CN (All 3 modalities) & \textbf{.7995 $\pm$ .0450} \\ \hline 
\end{tabular}
\caption{Performance comparison between Consensus Networks and fully-connected neural network classifiers having certain modality information.
\label{tab:fc-mods}}
\end{table}

\subsection{Dividing features by their natural modalities is desirable}
This is the second ablation study towards modality arrangement. We show that dividing features into subsets according to their natural modalities (i.e., the categories in which they are defined) is better than doing so randomly. 

In this experiment, we train CNs on features grouped by either their natural modalities, or randomly divided. For natural groupings, we try to let each group contain comparable number of features, resulting in the following settings:
\begin{itemize}
\item Two groups, natural: (a) acoustic + semantic, 216 features; (b) syntactic + PoS, 197 features.
\item Three groups, natural: (a) acoustic, 185 features; (b) semantic and PoS, 111 features; and (c) syntactic, 117 features. This is the default configuration used in other experiments in this paper.
\end{itemize}

For random grouping, we divide the features into almost equal-sized 2/3/4 groups randomly.
As shown in Table \ref{tab:mod-div}. The two natural modality division methods produce higher accuracies than those produced by any of the random modality division methods.

\begin{table}[h]
\centering 
\begin{tabular}{|l c|}
\hline 
Modality division method & Accuracy \\ \hline 
Three groups, random & .7408 $\pm$ .0340 \\
Two groups, random & .7623 $\pm$ .0164 \\
Four groups, random & .7666 $\pm$ .0141 \\ 
Two groups, natural & .7769 $\pm$ .0449 \\
Three groups, natural & \textbf{.7995 $\pm$ .0450} \\ \hline 
\end{tabular}
\caption{Performance comparison between different modality division methods, sorted by accuracy.}
\label{tab:mod-div}
\end{table}

\begin{figure*}  
\begin{subfigure}[b]{.19\textwidth}
	\centering 
    \includegraphics[scale=0.1]{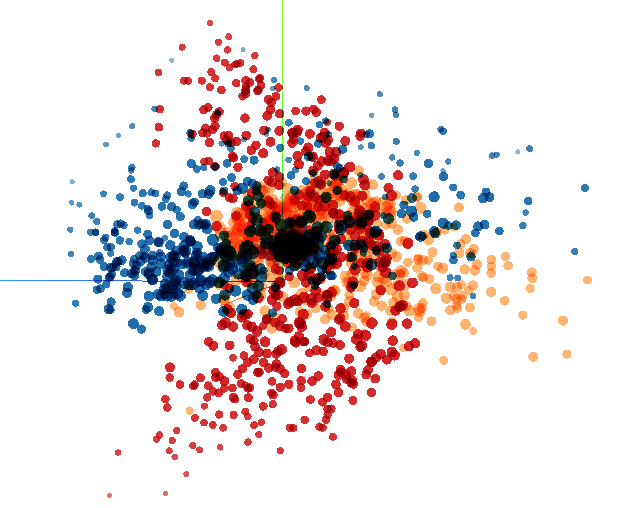}
    \caption{Step 5 \newline 
    $\mathcal{L_D}=1.34$ \newline
    Val accr $.74\%$ \newline 
    Variance $71.2\%$}
\end{subfigure}
\begin{subfigure}[b]{.19\textwidth}
	\centering 
    \includegraphics[scale=0.1]{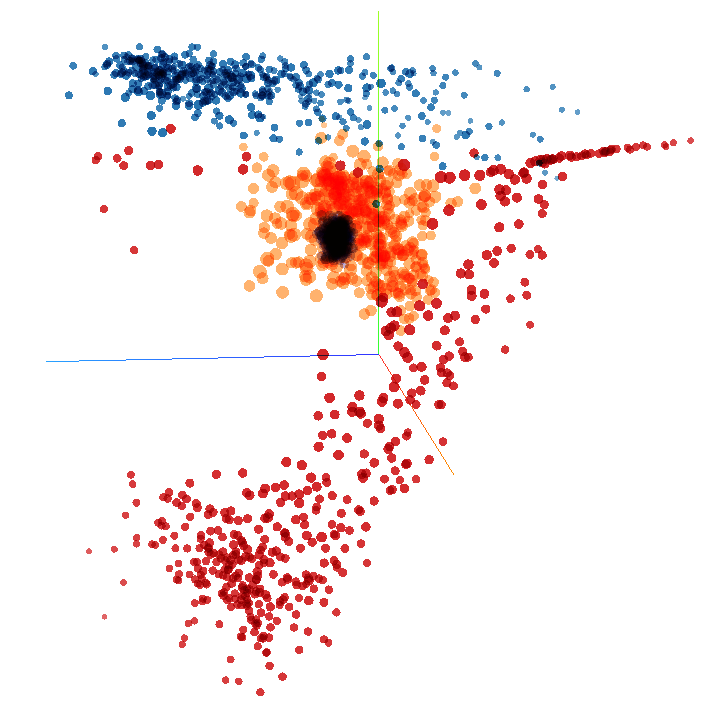}
    \caption{Step 10 \newline 
    $\mathcal{L_D}=1.32$ \newline 
    Val accr $.76\%$ \newline 
    Variance $72.9\%$}
\end{subfigure}
\begin{subfigure}[b]{.19\textwidth}
	\centering 
    \includegraphics[scale=0.1]{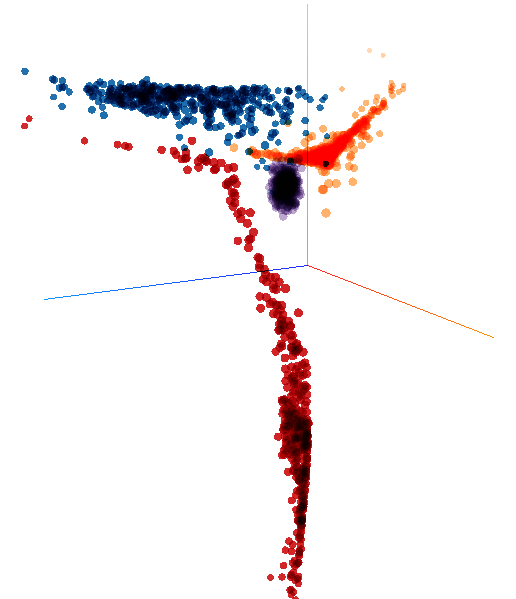}
    \caption{Step 20 \newline 
    $\mathcal{L_D}=1.17$ \newline 
    Val accr $.77\%$ \newline 
    Variance $77.1\%$}
\end{subfigure}
\begin{subfigure}[b]{.19\textwidth}
	\centering 
    \includegraphics[scale=0.1]{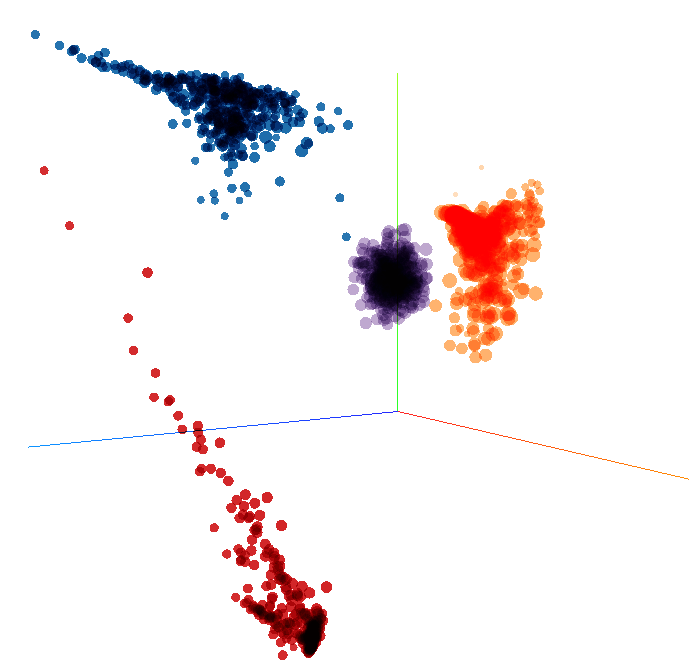}
    \caption{Step 30 \newline 
    $\mathcal{L_D}=0.89$ \newline 
    Val accr $.79\%$ \newline 
    Variance $77.0\%$}
\end{subfigure}
\begin{subfigure}[b]{.19\textwidth}
	\centering 
    \includegraphics[scale=0.1]{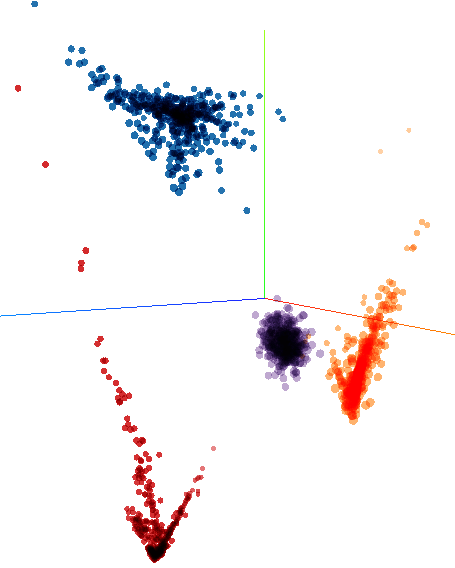}
    \caption{Step 40\newline 
    $\mathcal{L_D}=0.64$ \newline 
    Val accr $.79\%$ \newline 
    Variance $80.8\%$}
\end{subfigure}
\caption{Initially, the representations from the three modalities are mixed. As the training go on, the three modalities gradually form three symmetric ``petals'', while the noise Gaussian modality stays in the center. These petals do not overlap, as they contain complementary information when combined and passed into the classifier. Instead, their distributions become symmetric.
\label{fig:pca-interpretations}}
\end{figure*}

\subsection{Visualizing the representations}
\label{subsec:visualize}
To further understand what happens inside consensus network models during training, we visualize the representation vectors with PCA. Figure \ref{fig:pca-interpretations} consists of the visualizations drawn from an arbitrary trial in training the model. Each representation vector is shown on the figure as a data point, with its color representing its originating modality  (including the noise modality).

Several common themes could be observed:
\begin{enumerate}
	\item {\em The clusters are symmetric}. Initially the configurations of representations largely depend on the initializations of the network parameters. Gradually the representations of the same modality tend to form clusters. Optimizing the ePhysicians towards both targets make they compress modalities into representations which are symmetrical in an {\em aggregate} manner. 
	
	\item {\em The agreements are simple}. The variances explained by the first a few principal components usually increase as the optimizations proceed. When distilling information relevant to detection, the agreement tend to become simple.
	
	\item {\em The agreements are imperfect}. As shown in Figure \ref{fig:pca-interpretations}, the modal representations do not overlap. Also, the discriminator loss is low (usually at $10^{-3}$ when training is done). This confirms that these representations are still easily distinguishable. This may because the modalities inherently have some {\em complementary} information, leading to the ePhysicians projecting the modalities differently.
    
    \item {\em The representations are complex}. Their shapes do not resemble the noise representations (Gaussian) lying at the center of the three petals. This shows that the representations are not simply Gaussian.
    
    \item {\em The accuracy increases}. The accuracy in validation set generally increases as the training proceeds. Note that the distributions of representation vectors are increasingly similar in shape but remains distinct in spatial allocations. This confirms our conjecture that the information about cognitive impairment resides in complicated details instead of superficial statistics, which neural networks could represent.
\end{enumerate}

\subsection{Evaluation against benchmark algorithms}
With the previous sets of experiments, we have a best working architecture. We now evaluate it against traditional classifiers, which are used by the state-of-the-art papers \citep{hernandez2018computer,santos2017enriching,fraser15-JAD} on our 413 features. Note that the results could be different from what they reported, because the feature sets are different.

We test several traditional supervised learning benchmark algorithms here: support vector machine (SVM), quadratic discriminant analysis (QDA), random forest (RF), and Gaussian process (GP). For completeness, multiple layer perceptrons (MLPs) containing all features as inputs are also mentioned in Table \ref{tab:classifiers}. On the binary classification task (healthy control vs. dementia), our model does better than them all.

\begin{table}[h]
\centering 
\begin{tabular}{|l c c|}
\hline 
Classifier & Micro F1 & Macro F1  \\ \hline 
SVM & .4810 $\pm$ .0383 & .6488 $\pm$ .0329 \\ 
QDA & .5243 $\pm$ .0886 & .5147 $\pm$ .0904 \\ 
RF & .6184 $\pm$ .0400 & .6202 $\pm$ .0422\\  
GP & .6775 $\pm$ .0892 & .6873 $\pm$ .0819 \\  
MLP & .7528 $\pm$ .0520 & .7561 $\pm$ .0444 \\  
CN & \textbf{.7995 $\pm$ .0450} & \textbf{.7998 $\pm$ .0449} \\ \hline 
\end{tabular}
\caption{Comparison with different traditional classifiers in AD:Control classification task. In particular, our model has higher accuracy than the best traditional classifier, MLP.
\label{tab:classifiers}}
\end{table}

\section{Conclusion and future works}
We introduce the consensus network framework, in which neural networks are encouraged to compress various modalities into indistinguishable representations (`interpretation vectors'). 
We show that consensus networks, with the noise modality and cooperative optimization, improve upon traditional neural network baselines given the same features. 
Specifically, with all 413 linguistic features, our models outperform fully-connected neural networks and other traditional classifiers used by state-of-the-art papers.

In the future, the ``agreement among modalities'' concept may be applied to design objective functions for training classifiers in various tasks, and from other data sets (for example, education and occupation ``modalities'' for the bank marketing prediction task).
Furthermore, the mechanisms that represent linguistic features into symmetric spaces should be analyzed within the context of explainable AI.

\bibliographystyle{acl_natbib}
\bibliography{bibliography}

\newpage
\section*{Appendix}
\subsection*{Linguistic Features}
\paragraph{Acoustic}
\begin{itemize}
	\item The fluency of speech. We quantify it with phonation rate, duration of pauses, and number of filled pauses (e.g., \emph{"um"}) of various lengths.
    \item Following the convention of speech processing literatures \citep{Zhou2016,yancheva-fraser-rudzicz:SLPAT2015,Zhao2014}, we compute Mel-scaled cepstral coefficients (MFCCs) containing the amount of energy in 12 different frequency intervals for each time frame of 40 milliseconds, as well as their first- and second-order derivatives. We calculate the mean, variance, kurtosis, and skewness of the MFCCs and include them as features.
\end{itemize}

\paragraph{Semantic and Lexical}
\begin{itemize}
	\item Lexical norms, including age-of-acquisition, familiarity, imageability, and frequency \citep{taler2009comprehension}. These are averaged over the entire transcript and specific PoS categories, respectively.
    \item Lexical richness, including moving-average type-token ratio over different window sizes \citep{covington2010cutting}, Brunet's index, and Honor\'e's statistics \citep{guinn2012language}.
    \item Cosine similarity statistics (minimum, maximum, average, etc.) between pairs of utterances (represented as sparse vectors based on lemmatized words)
    \item Average word length, counts of total words, not-in-dictionary words, and fillers. The dictionary we use contains around 98,000 entries, including common words, plural forms of countable nouns, possessive forms of subjective nouns, different tenses of verbs, etc.
\end{itemize}

\paragraph{Syntactic}
\begin{itemize}
	\item Composition of languages. We describe it by several features, including the average proportion of context-free grammar (CFG) phrase types\footnote{number of words in these types of phrases, divided by the total number of words in the transcript}, the rates of these phrase types\footnote{number of occurrences in a transcript, divided by the total number of words in the transcript}, and the average phrase type length\footnote{number of words belonging to this phrase type in a transcript, divided by the occurrences of this phrase type in a transcript} \cite{chae2009predicting}
    \item The syntactic complexity of languages. We characterize it by the average heights of the context-free grammar (CFG) parse trees, across all utterances in each transcript. Each tree comes from an utterance parsed by a context free grammar parser (\texttt{LexParser} implemented in Stanford CoreNLP \cite{corenlp}). In addition, we compute the Yngve scores statistics of CFG parse trees \citep{yngve1960model,roark2007syntactic}, where Yngve score is the degree of left-branching of each node in a parsed tree.
    \item Syntactic components. We describe them by including the number of occurrences of a set of 104 context-free production rules (e.g.,\texttt{S->VP}) in the CFG parse trees.
\end{itemize}

\paragraph{Part-of-speech}
\begin{itemize}
	\item The number of occurrences of part-of-speech (PoS) tags from Penn-treebank\footnote{Using \url{https://spacy.io}}.
    \item The ratio of occurrences of several PoS tags, including noun-pronoun ratio.
    \item Number of occurrences of words in each of the five categories: subordinate (e.g: ``because", ``since", etc.), demonstratives (e.g: ``this", ``that"), function (e.g: words with PoS tag CC, DT, and IN), light verbs (e.g: ``be", ``have"), and inflected verbs (words with PoS tag VBD, VBG, VBN, and VBZ), borrowing the categorization method in \citet{kortmann2004global}
\end{itemize}

\end{document}